\documentclass[runningheads]{llncs}

\usepackage{eccv}

\usepackage{eccvabbrv}

\usepackage{graphicx}
\usepackage{booktabs}
\usepackage{xcolor}
\usepackage{adjustbox}
\usepackage[table]{xcolor}
\usepackage{float}

\usepackage{tabularx}
\usepackage{booktabs}

\usepackage[accsupp]{axessibility}  

\usepackage[pagebackref,breaklinks,colorlinks,citecolor=eccvblue]{hyperref}

\usepackage{orcidlink}
\usepackage{marvosym}

\begin{document}

\title{NoDrift3R: Raymap-Guided Coupling for Drift-Robust Unposed Feed-Forward 3D Reconstruction} 

\titlerunning{NoDrift3R}


\author{Xiangyu Sun\inst{1}\thanks{Intern at Horizon Robotics}\orcidlink{0009-0009-0625-4240} \and
Liu Liu\inst{2}\textsuperscript{\Letter} \and
Seungkwon Yang\inst{3} \and
Jingbing Han\inst{2} \and
Seungtae Nam\inst{3} \and
Zhizhong Su\inst{2}\orcidlink{0000-0003-2312-9985} \and
Eunbyung Park\inst{3}\textsuperscript{\Letter}\orcidlink{0000-0003-4071-2814}}

\authorrunning{X. Sun et al.}

\institute{Sungkyunkwan University, South Korea \and
Horizon Robotics, China \and
Yonsei University, South Korea \\
\vspace{2.0mm}
\url{https://xiangyu1sun.github.io/NoDrift3R-project-page/}}


\maketitle

\begingroup
\renewcommand\thefootnote{}
\footnotetext{\textsuperscript{\Letter} Corresponding authors.}
\addtocounter{footnote}{-1}
\endgroup

\begin{figure*}[t]
  \centering
  \includegraphics[width=\textwidth]{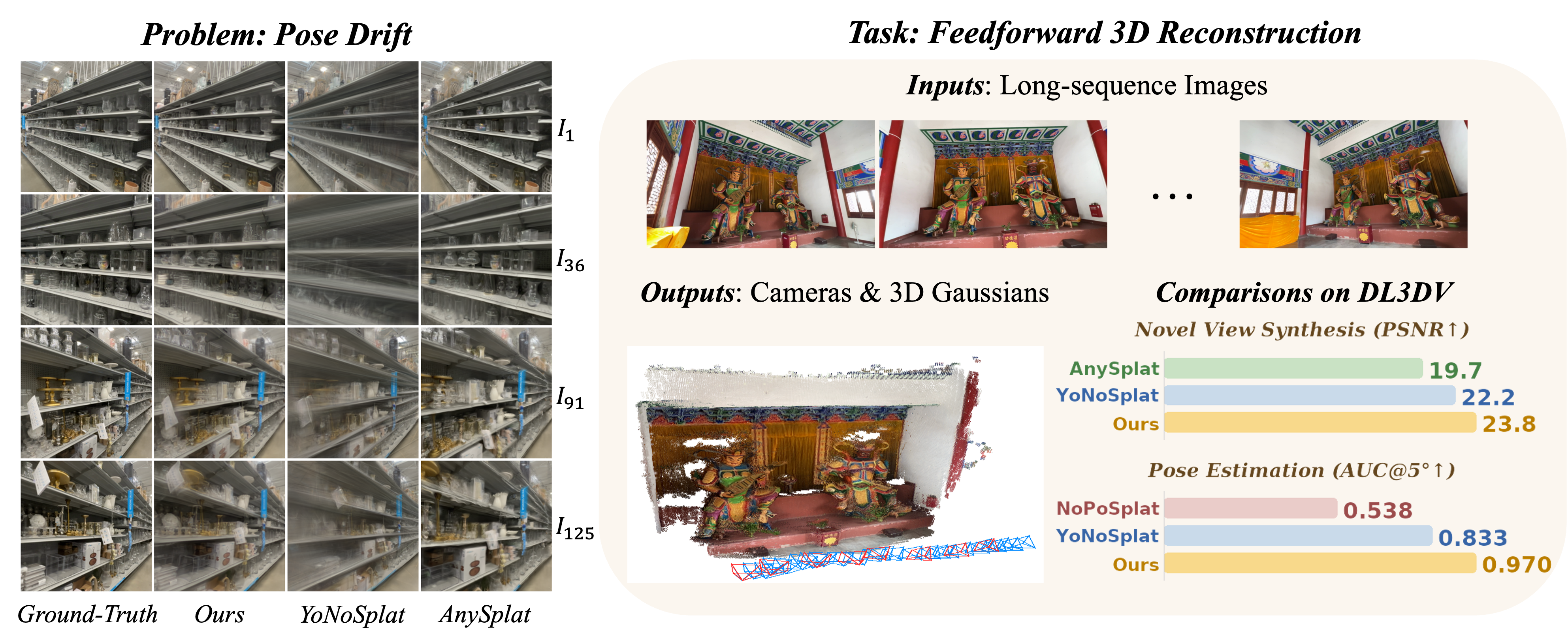}

  
  \captionof{figure} {Overview of our synergistic pose-free framework for feed-forward 3D reconstruction. Our method effectively suppresses the pose drift problem, especially in long-sequence settings. Left: representative failure cases of existing methods under pose drift. Right: our pipeline and outputs (camera poses and 3D Gaussians).}
  \label{fig:problem_define}

  
\end{figure*}

\vspace{-3.0mm}

\begin{abstract}

Pose-Free Feed-forward 3D Gaussian Splatting (3DGS) has recently emerged as a powerful paradigm for fast scene reconstruction. However, its performance degrades significantly in long image sequences due to cumulative camera pose estimation drift, which propagates errors into geometric modeling and severely limits rendering fidelity. In this work, we revisit the long-sequence bottleneck and identify pose drift as the primary factor restricting reconstruction quality. Furthermore, while SfM-based pseudo ground-truth poses introduce sensor noise, purely rendering-based supervision often leads to optimization instability and local minima due to the entangled optimization of geometry and pose.
To address the challenges, we propose a synergistic pose-free framework that explicitly couples geometry and appearance via a Raymap-Guided Coupling Module (RGC). Concretely, we anchor Gaussian centers to raymap-induced geometry and jointly optimize RGB reconstruction, raymap consistency, and camera regularization under a unified objective, yielding a bidirectional feedback loop: stronger geometry improves rendering, and appearance supervision in turn refines geometry and pose. To further stabilize learning across wide temporal ranges, we introduce a Dual-Frequency Viewpoint Scheduling strategy that combines easy-to-hard interval expansion with replay of short-interval pairs. 
Extensive experiments across in-domain and cross-domain datasets show consistent gains in both rendering and pose estimation, with notably improved robustness on long sequences. Ablation studies validate our central insight: explicitly designed geometry-appearance synergy is the key to scalable and drift-robust pose-free feed-forward 3D reconstruction. 


\end{abstract}

\section{Introduction}
\label{sec:intro}
Feed-forward 3D Gaussian Splatting (3DGS) has recently emerged as a powerful paradigm for rapid 3D reconstruction and novel view synthesis by predicting scene representations directly from multi-view images. Compared to conventional per-scene optimization methods~\cite{mildenhall2021nerf, kerbl20233dgs}, feed-forward approaches offer significantly faster reconstruction times, opening the door to applications that require fast 3D reconstruction in various practical scenarios. Early feed-forward pipelines typically assume known camera poses and focus on predicting Gaussian representations conditioned on calibrated multi-view observations~\cite{charatan2024pixelsplat, chen2024mvsplat, xu2025depthsplat, kang2025ilrm, nam2025generative}. More recently, pose-free feed-forward models have been proposed to jointly infer camera geometry and scene representation directly from unposed image sequences~\cite{ye2024noposplat, sun2025uni3r, jiang2025anysplat, ye2025yonosplat, kang2025selfsplat}, thereby expanding the applicability of feed-forward 3DGS to settings where camera poses are not available in inference.

Despite encouraging progress, pose-free feed-forward 3DGS remains challenging, particularly for long sequences or wide-baseline viewpoints. A central difficulty arises from the strong coupling between predicted camera parameters and scene representation: errors in pose estimation can directly distort the reconstructed representation, while inaccuracies in scene structure can further degrade pose predictions. This feedback loop often leads to accumulated pose drift, where small misalignments gradually amplify over time and deteriorate reconstruction quality. As a result, achieving stable and globally consistent reconstruction from long, unposed image sequences remains an open challenge.

Furthermore, we observe that training instability is a pervasive challenge inherent to all pose-free paradigms, transcending the choice of model initialization. Even when fine-tuning atop powerful 3D foundation models—such as Uni3R~\cite{sun2025uni3r} with VGGT~\cite{wang2025vggt} or YoNoSplat~\cite{ye2025yonosplat} with Pi3~\cite{wang2025pi3}—optimization remains volatile. 
While YoNoSplat attempts to stabilize this via a teacher-forcing mix-training strategy, and E-RayZer~\cite{zhao2025erayzer} introduces a curriculum learning approach for training from scratch, these methods still struggle to balance stability with fine-grained precision.

Specifically, we identify a critical limitation in E-RayZer’s linear visual-overlap extension: it tends to degrade performance at small-interval samples, which are essential for maintaining local geometric consistency. To bridge this gap, we design a Dual-Frequency Viewpoint Scheduling. By scheduling samples from "easy" (high overlap) to "hard" (low overlap) while systematically re-exposing the model to small-interval cases, our curriculum stabilizes the optimization process across all stages. This approach ensures robust performance across arbitrary frame intervals, effectively suppressing drift and maintaining high-fidelity reconstruction even in long, complex sequences.

\begin{figure}[t]
  \centering
  \includegraphics[width=0.9\textwidth]{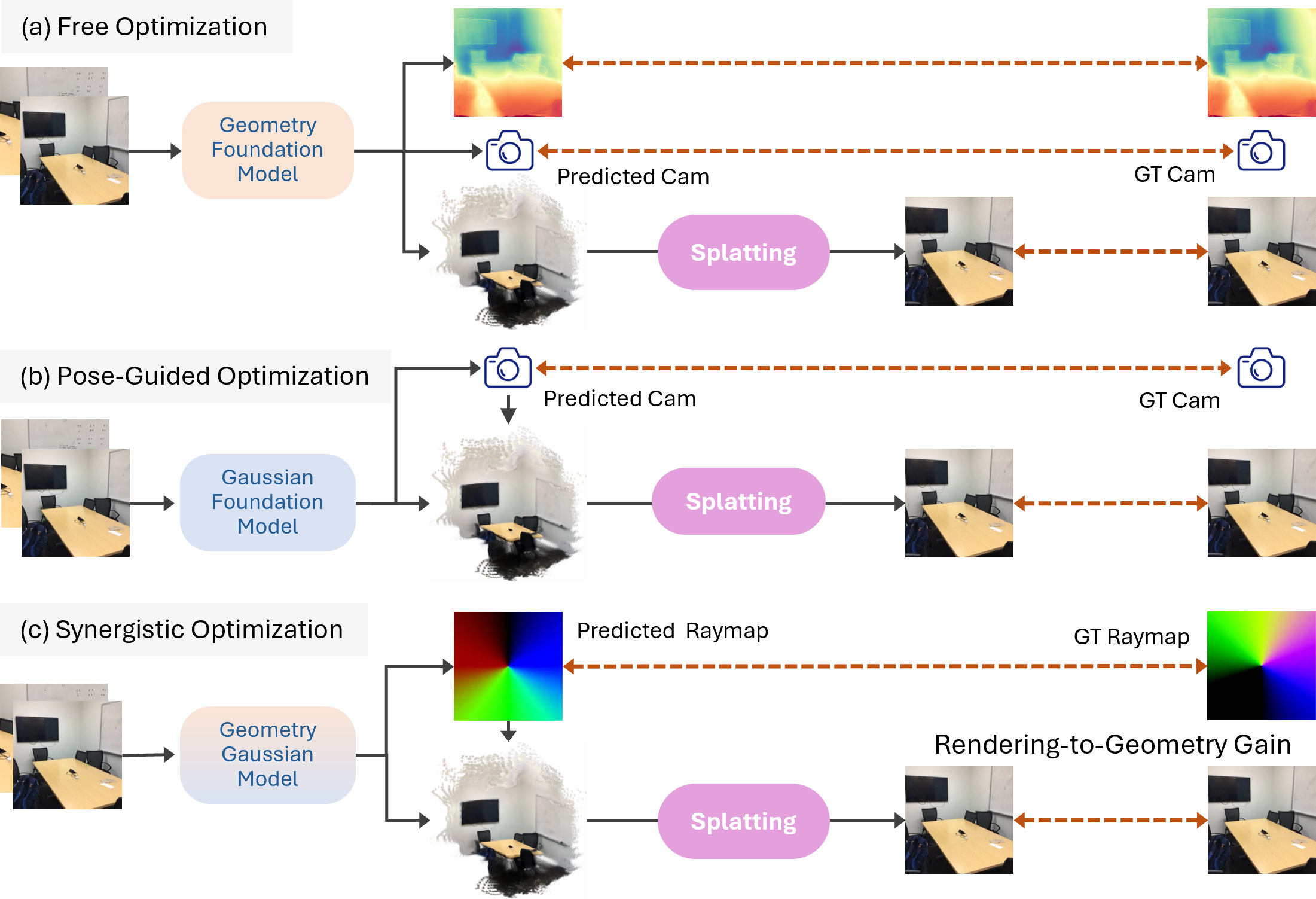}
  \caption{Our synergistic framework for "Rendering-to-Geometry Gain".}
  \label{fig:problem_formulation}
\end{figure}

In this work, we propose a synergistic joint optimization framework (\cref{fig:problem_formulation}-(c)) that tightly couples camera geometry and appearance generation through a Raymap-Guided Coupling (RGC) module. Existing approaches generally follow two design paradigms. The first (\cref{fig:problem_formulation}-(a)) predicts pixel-aligned Gaussian representations directly from images without explicitly leveraging estimated camera poses~\cite{jiang2025anysplat, sun2025uni3r}. While this strategy can produce reasonable renderings, it lacks a mechanism to tightly connect pose estimation with the resulting 3D representation. The second paradigm (\cref{fig:problem_formulation}-(b)) conditions Gaussian prediction on estimated camera poses, allowing geometry and appearance to interact to some extent~\cite{ye2025yonosplat}. However, this interaction remains limited because camera poses are represented by low-dimensional parameters (e.g., rotation and translation), creating an information bottleneck that cannot convey the dense geometric cues required for accurate Gaussian placement. In contrast, our RGC module establishes a tighter interaction between camera pose estimation and Gaussian generation by predicting dense raymaps that directly determine the spatial distribution of Gaussians. Because raymaps encode per-pixel ray directions, both appearance supervision and geometric constraints jointly influence every per-pixel Gaussian representation during training. This design forms a bidirectional optimization loop in which rendering supervision refines geometry while geometric predictions regularize appearance synthesis. As a result, our framework enables more stable pose-free optimization and improved reconstruction quality across long sequences and wide view intervals.



In sum, our contributions are threefold:
\begin{itemize}
\item Identification of the Long-Sequence Bottleneck: We provide a comprehensive analysis of feed-forward 3D Gaussian Splatting and identify that cumulative pose estimation drift is the primary bottleneck limiting rendering quality in long sequences. We further demonstrate that while SfM~\cite{schonberger2016sfm}-based pseudo-GT poses introduce sensor noise, pure rendering-based supervision (e.g., NoPoSplat~\cite{ye2024noposplat}, Uni3R~\cite{sun2025uni3r}) often leads to optimization instability and local optima due to the simultaneous optimization of geometry and pose.

\item Dual-Frequency Viewpoint Scheduling: We propose a novel training schedule driven by visual overlap and a replay mechanism. By systematically re-exposing the model to high-overlap samples while progressively introducing low-overlap challenges, our schedule stabilizes the optimization of coupled geometry and pose. This ensures robust, drift-free reconstruction across arbitrary intervals and wide-range viewpoints.

\item A Synergistic Framework for "Rendering-to-Geometry Gain": We propose a novel pose-free framework that integrates 3DGS rendering supervision with explicit raymap constraints. Unlike previous methods that optimize geometry and pose in isolation, our approach establishes a positive feedback loop—which we term "Rendering-to-Geometry Gain"—where appearance consistency serves as a powerful geometric prior. This synergy effectively resolves parameter ambiguity and suppresses drift in extended sequences, allowing our model to achieve state-of-the-art (SOTA) performance in both rendering fidelity and camera pose accuracy.
\end{itemize}




\section{Related Work}
\noindent {\bf Pose-dependent Feed-forward 3D Reconstruction.} Traditional 3D reconstruction and neural rendering have been significantly advanced by approaches such as NeRF~\cite{mildenhall2021nerf} and 3D Gaussian Splatting (3DGS)~\cite{kerbl20233dgs}. However, these methods typically rely on computationally intensive per-scene optimization over dense image sets, which limits their scalability and practicality in real-world applications. To overcome this limitation, feed-forward 3D reconstruction models have been proposed, aiming to predict 3D representations in a single forward pass by learning generalizable priors from large-scale datasets. Early feed-forward approaches primarily focused on multi-view stereo paradigms, leveraging carefully designed architectures that incorporate geometric reasoning mechanisms such as cost volumes and epipolar geometry constraints~\cite{chen2021mvsnerf, chen2024mvsplat, chen2024mvsplat360, xu2025depthsplat, charatan2024pixelsplat, nam2025generative}. More recently, architectures based on Vision Transformers~\cite{dosovitskiy2020image} that do not rely on such geometric constraints have emerged as a unified framework for multi-view geometry modeling~\cite{hong2023lrm, zhang2024gs, imtiaz2025lvt, ziwen2025long, kang2025ilrm, wang2026tttlrm, kang2025multi, jin2024lvsm, dens3r}. To address long-sequence reasoning, tttLRM~\cite{wang2026tttlrm} adopts Test-Time Training~\cite{sun2024learning, zhang2025test} layers, while Long-LRM~\cite{ziwen2025long} incorporates Mamba2~\cite{dao2024transformers} blocks for efficient sequence modeling. Meanwhile, iLRM~\cite{kang2025ilrm} and MVP~\cite{kang2025multi} carefully design the use of image or viewpoint embeddings to enhance consistency and scalability. 

\noindent {\bf Pose-free Feed-forward 3D Reconstruction.} While posed models deliver high fidelity, a fundamental limitation of these approaches is their strong dependence on accurate camera pose, which is usually obtained via Structure from Motion (SfM) methods like COLMAP~\cite{schonberger2016sfm}. In unconstrained real-world scenarios, obtaining precise camera parameters is challenging or infeasible, which substantially restricts their practical applicability. While early methods~\cite{lin2021barf, wang2021nerfmm, meng2021gnerf, fu2024colmap, fan2024instantsplat} address this by jointly optimizing poses alongside scene representations, pose-free feed-forward models further bypass per-scene optimization entirely. They directly infer 3D scene representations or camera parameters from sparse inputs in a single pass, operating either on uncalibrated images or with known intrinsics. Recent frameworks bypass traditional SfM pipelines by anchoring 3D primitives directly within a canonical coordinate space~\cite{jiang2025anysplat, ye2024noposplat, hong2024pf3plat, zhao2025erayzer, huang2025no}, by consolidating local per-view geometries into a shared global coordinate frame~\cite{zhang2025flare, kang2025selfsplat, deng2025selfi, ye2025yonosplat}, or by fusing multi-view features into a view-agnostic latent representation for direct global prediction~\cite{sun2025uni3r}. Despite their efficiency, the inherent coupling between predicted camera parameters and scene geometry frequently leads to accumulated pose drift. In contrast, our model incorporates a raymap loss that allows for the joint optimization of geometry and appearance, effectively suppressing drift accumulation across extended trajectories.

\noindent {\bf Multi-view Visual Geometry Foundation Models.} SfM~\cite{schonberger2016sfm} and Multi-View Stereo remain widely used to obtain reliable visual geometry, but their multi-stage pipelines are notoriously tedious and computationally cumbersome. Recently, the landscape of multi-view visual geometry has been reshaped by large-scale foundation models. DUSt3R~\cite{wang2024dust3r} and MASt3R~\cite{leroy2024grounding} reformulate the 3D reconstruction task as a pairwise dense pointmap regression problem, bypassing the need for prior camera calibration. 
VGGT~\cite{wang2025vggt} employs an alternating frame-wise and global self-attention mechanism to jointly predict camera parameters, keypoint tracking, depth, and pointmaps from multiple views in a single pass, avoiding the pairwise limitations.
$\pi^3$~\cite{wang2025pi3} improves VGGT by eliminating the inductive bias of anchoring scenes to a fixed reference frame by leveraging a fully permutation-equivariant architecture to predict affine-invariant outputs. Concurrently, frameworks introduce flexible architectures capable of seamlessly ingesting available geometric priors to enhance global alignment; for instance, MapAnything~\cite{keetha2025mapanything} can optionally condition on known camera intrinsics, poses, and sparse depth, while Depth Anything v3~\cite{lin2025depthanythingv3} similarly supports the integration of camera intrinsics and poses. Because these models bypass traditional optimization and are considered to produce high-quality 3D features, they frequently serve as the foundational backbones or pseudo-label teachers for recent pose-free, feed-forward 3D reconstruction models.

\section{Method}

In this section, we first propose our Raymap-Guided Coupling Module (\cref{sec:raymap}). Then we elaborate on the Synergistic Framework (\cref{sec:loss}). Subsequently, we propose the Dual-Frequency Viewpoint Scheduling to train the model under both easy- and hard-overlap circumstances (\cref{sec:replay}).

\begin{figure*}[t]
  \centering
  \includegraphics[width=\textwidth]{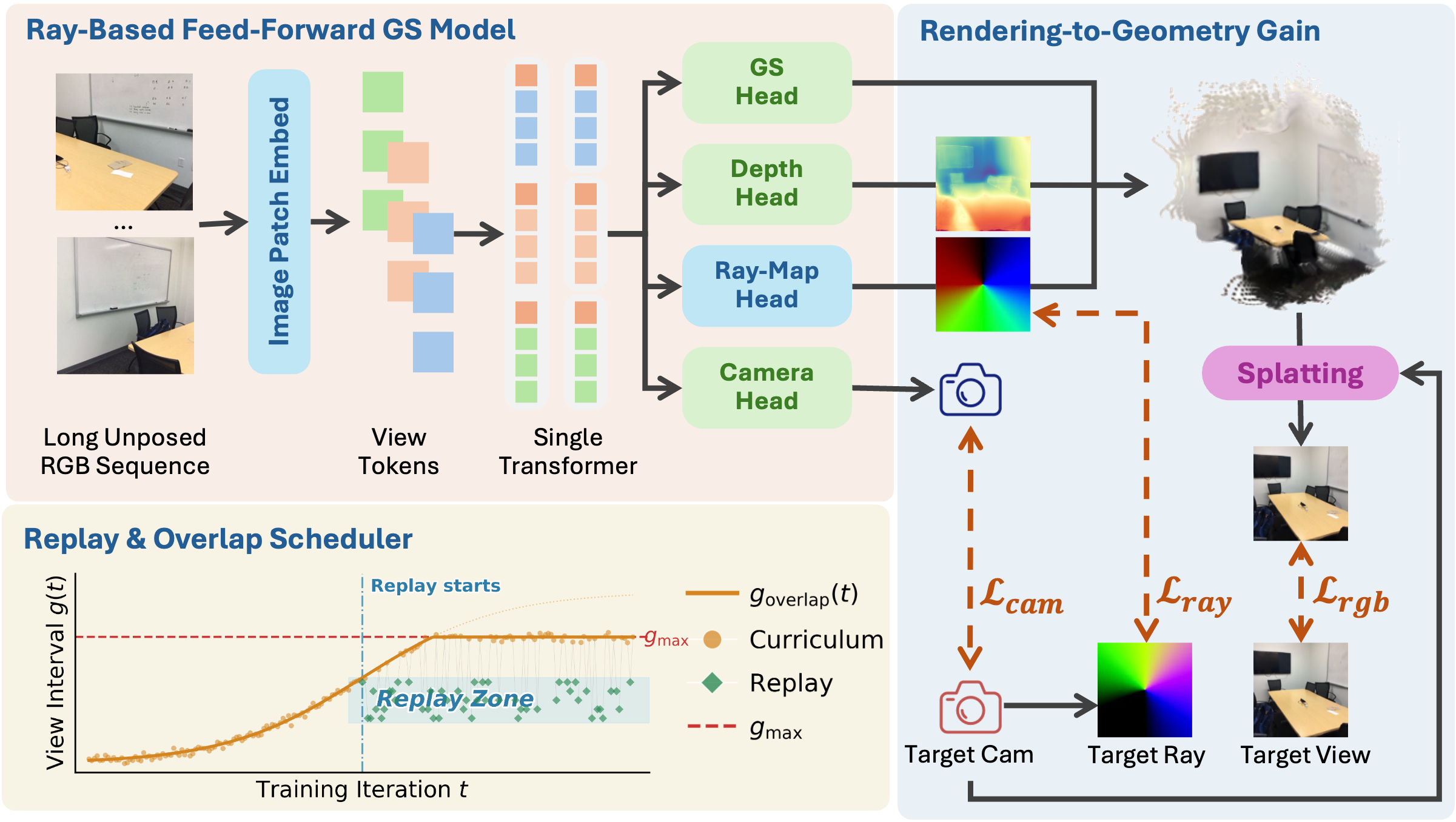}

  
  \caption{\textbf{Overview of the Synergistic Framework.} (a) Ray-Based Feed-Forward GS Model. We employ a single transformer (Vallina DINOv2 model), followed by Gaussian Head, Depth Head, Ray-Map Head, and Camera Head. (b) A Raymap-Guided Coupling Module for "Rendering-to-Geometry Gain". (c) Our Replay \& Overlap Scheduler leads to robust performance across arbitrary intervals.}
  \label{fig:pipeline}

  
\end{figure*}

\subsection{Raymap-Guided Coupling Module}
\label{sec:raymap}

Given an input image sequence $\{I_i\}_{i=1}^{N}$, we first apply a lightweight image patch-embedding to each view and obtain per-view tokens. These view tokens are then processed by a single multi-view transformer to perform inter-view interaction and produce latent tokens for each view. (e.g., Depth Anything v3~\cite{lin2025depthanythingv3}).

From these per-view latent tokens, a Dual-DPT~\cite{ranftl2021vision} head jointly predicts two geometric signals per view: a ray map $R \in \mathbb{R}^{HW \times 6}$ and a depth map $D \in \mathbb{R}^{HW}$. For each pixel, the first three channels of $R$ encode the ray direction (denoted by ${r} \in \mathbb{R}^{HW \times 3}$) and the last three channels encode the ray origin, ${o} \in \mathbb{R}^{HW \times 3}$. The depth map provides the scalar distance along each ray. We then lift pixels to 3D and obtain Gaussian centers using both signals:
\begin{equation}
  {p}_{j} = {o}_{j} + D_{j} \cdot {r}_{j},
\end{equation}
where ${r}_{j} \in \mathbb{R}^3$ and ${o}_{j} \in \mathbb{R}^3$ are the direction and origin at the pixel location $j$, and ${p}_{j} \in \mathbb{R}^{3}$ represents the corresponding Gaussian center position.
This raymap-guided lifting explicitly ties Gaussian positions to predicted geometry and serves as the geometric anchor for subsequent rendering and optimization.

After obtaining $D$ and $R$, we feed the corresponding latent features (i.e., the per-view latent tokens output by the single transformer) to the Gaussian head to predict Gaussian attributes. Following common 3DGS parameterizations,
each Gaussian is defined by its position and a full set of attributes. For each pixel-aligned primitive, our Gaussian head predicts $\{\alpha_j, {c}_j, {s}_j, {q}_j\}$,
where $\alpha_j \in [0,1]$ is opacity, $\mathbf{c}_j \in \mathbb{R}^3$ is color, $\mathbf{s}_j \in \mathbb{R}^3$ is anisotropic scale, and $\mathbf{q}_j \in \mathbb{R}^4$ is a unit rotation quaternion. Importantly, we used the lifted 3D position as the Gaussian center, $\mu_j={p}_j$.
To keep parameters in valid ranges, we apply
\begin{equation}
  \alpha_j = \sigma(f_j^{\alpha}), \quad {s}_j = \exp(f_j^{s}), \quad {q}_j = \mathrm{normalize}(f_j^{q}),
\end{equation}
where $f$ denotes the output from the Gaussian heads for each attribute, $\sigma(\cdot)$ is the sigmoid function. This design yields a complete Gaussian representation aligned with our raymap-derived geometry.

\subsection{Learning a Synergistic 3D Representation}
\label{sec:loss}
A key aspect of our method is the synergistic optimization between appearance and geometry. Unlike prior pipelines (e.g., YoNoSplat\cite{ye2025yonosplat} and AnySplat\cite{jiang2025anysplat}) that treat rendering supervision and geometric constraints as loosely coupled signals, our framework ties them together through raymap-derived Gaussian positions.
This coupling allows improvements in one branch to directly benefit the other, creating a self-reinforcing training dynamic that is especially important for long sequences and pose-free settings.

Our training objective couples rendering, camera, and raymap supervision in a unified formulation. Given an image sequence $\{I_i\}_{i=1}^{N}$, we use an RGB reconstruction loss that combines MSE loss and perceptual LPIPS loss:
\begin{equation}
  \mathcal{L}_{\text{rgb}} = \frac{1}{N} \sum_{i=1}^{n} \left(\lambda_{\text{mse}} \| I_i - \hat{I}_i \|_2^2 + \lambda_{\text{lpips}} \, \mathrm{LPIPS}(I_i, \hat{I}_i)\right),
\end{equation}
where $\hat{I}_i$ denotes the rendered images.
For camera regularization, we supervise the camera parameters predicted by the camera head with Huber loss:
\begin{equation}
  \mathcal{L}_{\text{cam}} = \frac{1}{N} \sum_{i=1}^{N} \mathrm{Huber}(\hat{{\theta}}_i - {\theta}_i),
\end{equation}
where $\hat{\theta}_i, {\theta}_i$ represent predicted and ground truth camera parameters, repsectively.
We further apply raymap loss to the per-frame predicted raymaps:
\begin{equation}
  \mathcal{L}_{\text{ray}} = \frac{1}{N} \sum_{i=1}^{N} \mathcal{L}_{\text{abs}}(\hat{R}_i, R_i).
\end{equation}
The total objective is a weighted sum as follows:
\begin{equation}
  \mathcal{L} = \mathcal{L}_{\text{rgb}} + \lambda_{\text{cam}} \mathcal{L}_{\text{cam}} + \lambda_{\text{ray}} \mathcal{L}_{\text{ray}}.
\end{equation}
Crucially, these losses are synergistic rather than independent. Our Gaussian positions are derived from depth and raymap; consequently, the RGB loss $\mathcal{L}_\text{rgb}$ backpropagates through the rendering process into the raymaps, allowing appearance cues to refine geometry.
Meanwhile, the raymap loss $\mathcal{L}_\text{ray}$ directly constrains ray directions, and since raymaps define the Gaussian distribution (positions), improved geometry leads to better rendering quality. This mutual reinforcement between raymap supervision and RGB supervision under a joint optimization objective plays an important role in achieving rendering-to-geometry gain, despite the simplicity of the losses.

\noindent\textbf{Insights.} This bidirectional coupling from our Raymap-Guided Coupling (RGC) module is the key advantage of our framework: appearance supervision sharpens geometry, and raymap supervision regularizes the spatial distribution of Gaussians, which in turn improves appearance. As a result, we obtain stable optimization and stronger generalization across arbitrary view intervals without introducing complex auxiliary losses. The simplicity of our loss design is therefore a feature, not a limitation—because the synergy is baked into the representation and the training flow itself. More detailed empirical evidence is provided in the ablation studies (Table~\ref{tab:loss_ablation_all}).

\subsection{Dual-Frequency Viewpoint Scheduling}
\label{sec:replay}
We adopt a curriculum-based training strategy that gradually increases the difficulty of view pairs from high overlap to low overlap configurations, a commonly used paradigm in pose-free and self-supervised learning~\cite{zhao2025erayzer}.
Unlike existing approaches, we develop two mechanisms tailored to our framework: (i) a bounded overlap-aware curriculum, and (ii) a replay mechanism that repeatedly injects small-interval samples to provide stochastic local geometric regularization.


\paragraph{Sparsity Upper-Bound Clipping.} 
First, we leverage DINOv2~\cite{oquab2023dinov2} to obtain pairwise overlap scores~\cite{zhao2025erayzer} by calculating cosine similarity as follows:
\begin{equation}
  o(i,j)=\textrm{cos}(\textrm{DINOv2}(I_i), \textrm{DINOv2}(I_j)).
\end{equation}
Then, we calculate the overlap-aware view interval (the number of frames between input view images) at training iteration $t$, $g_\text{overlap}(t)$, is defined as,
\begin{equation}
  g_\text{overlap}(t) = \min\big(g_\text{curr}(t),\, g_\text{max}),
\end{equation}
where $g_\text{curr}(t)$ denotes the largest view intervals whose overlap score exceeds a target threshold at training iteration $t$. The target threshold is gradually annealed according to a schedule, decreasing from $1.0$ to $0.75$. In addition, we clip the maximum interval $g_\text{max}=15$ between input image pairs to avoid overly sparse supervision. This constraint keeps the scheduler within a bounded temporal range while exposing the model to progressively larger view intervals.

\paragraph{Stochastic Replay Regularization.} To prevent the performance degradation on small-interval evaluation, we add a replay branch that samples a large-overlap (small-interval) intervals with probability $p_{\text{small}}=0.5$ during the last half of the training phase. Specifically, we draw $g_{\text{small}} \sim \mathcal{U}\{6,\dots,10\}$,
and select it with probability $p_{\text{small}}$. More formally, we select the view interval at training iteration $t$, $g(t)$, as follows:
\begin{equation}
  g(t) =
  \begin{cases}
    g_{\text{small}}, & \text{with prob. } p_{\text{small}}, \\
    g_\text{overlap}(t), & \text{with prob. } 1- p_{\text{small}}.
  \end{cases}
\end{equation}
This replay curriculum preserves the benefits of an easy-to-hard schedule while ensuring the model repeatedly sees small-interval scenes, leading to robust performance across arbitrary intervals.

\section{Experiments}

\subsection{Experiments Setup}
\subsubsection{Implementation Details}
Our framework is implemented in PyTorch~\cite{paszke2019pytorch}. The model employs a standard ViT~\cite{dosovitskiy2020image} backbone (vanilla DINOv2~\cite{oquab2023dinov2}) without structural modifications to the transformer blocks. To evaluate the scalability and effectiveness of our framework, we adopt two variants of the Depth Anything V3~\cite{lin2025depthanythingv3} architecture as our backbone:
\begin{itemize}
\item \textbf{Giant Model:} For primary comparisons with state-of-the-art (SOTA) methods, we utilize the Giant variant. Built upon a backbone with a hidden dimension of 1536 and 40 transformer blocks, this high-capacity model provides precise initial geometry and raymap priors, maximizing reconstruction fidelity.
\item \textbf{Large Model:} For all ablation studies, we employ the Large variant to balance computational efficiency with rigorous evaluation. This version uses a backbone with a hidden dimension of 1024 and 24 transformer blocks.
\end{itemize}
By conducting ablations on the Large model, we demonstrate the Dual-Frequency Viewpoint Scheduling and Rendering-to-Geometry Gain yield consistent performance improvements regardless of the backbone scale. The Giant Model is trained on 16 H100 GPUs for 150k steps with a batch size of 1 for each, while the Large Model is trained on 8 RTX5090 GPUs for 150k steps with a batch size of 1 for each. Both are initialized from Depth Anything v3 pretrained weights.

\subsubsection{Metrics}
For quantitative results on the novel view synthesis, we evaluate with the commonly used metrics PSNR, SSIM~\cite{wang2004image}, and LPIPS~\cite{zhang2018unreasonable}. For pose estimation, we report the area under the cumulative angular pose error curve (AUC) thresholded at \(5^{\circ}\), \(10^{\circ}\), and \(20^{\circ}\). For a fair comparison, all experiments are conducted on 224$\times$224 resolutions.

\subsubsection{Datasets}
We conduct experiments on the DL3DV dataset~\cite{ling2024dl3dv}, which consists of large-scale multi-view scenes spanning diverse indoor and outdoor environments. Each scene is processed using COLMAP~\cite{schonberger2016sfm, schonberger2016mvs} to obtain accurate camera poses, containing approximately 250--350 frames per scene. We further train the model on the RealEstate10K (RE10K) dataset~\cite{zhou2018re10k}. For both in-domain and cross-domain evaluation on DL3DV, RE10K, and Scannet++~\cite{yeshwanth2023scannet++} dataset, we follow the same settings as YoNoSplat~\cite{ye2025yonosplat}.



\begin{figure}[t]
  \centering
  \includegraphics[width=\textwidth]{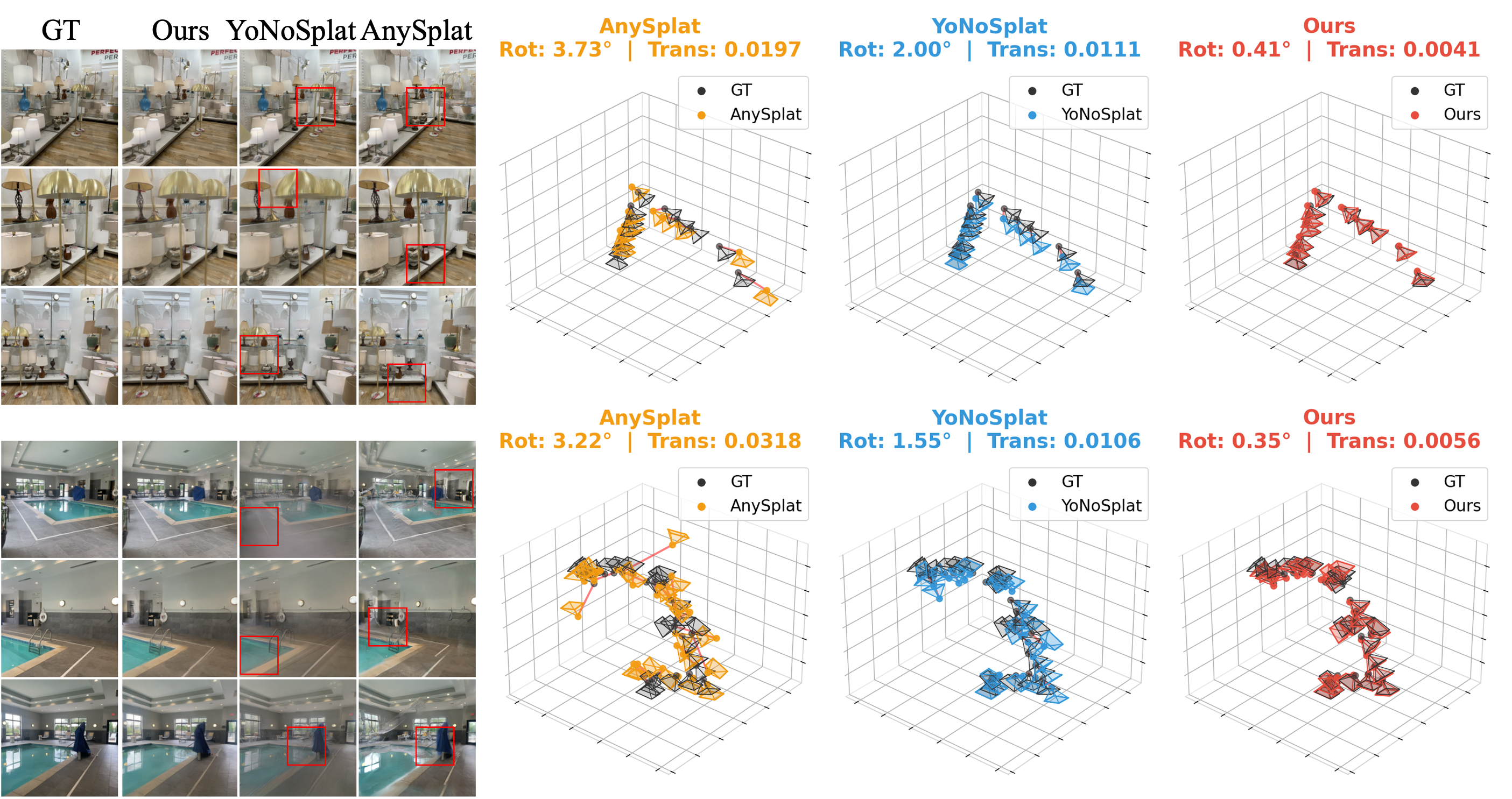}

  
  \caption{\textbf{Pose visualization and compared with representative methods.} Left: the rendering visualization of the target views. Right: we visualize the predicted context cameras, which shows our model can handle the pose drift problem.}
  \label{fig:dl3dv_pose}
  
\end{figure}

\begin{figure}[t]
  \centering
  \includegraphics[width=\textwidth]{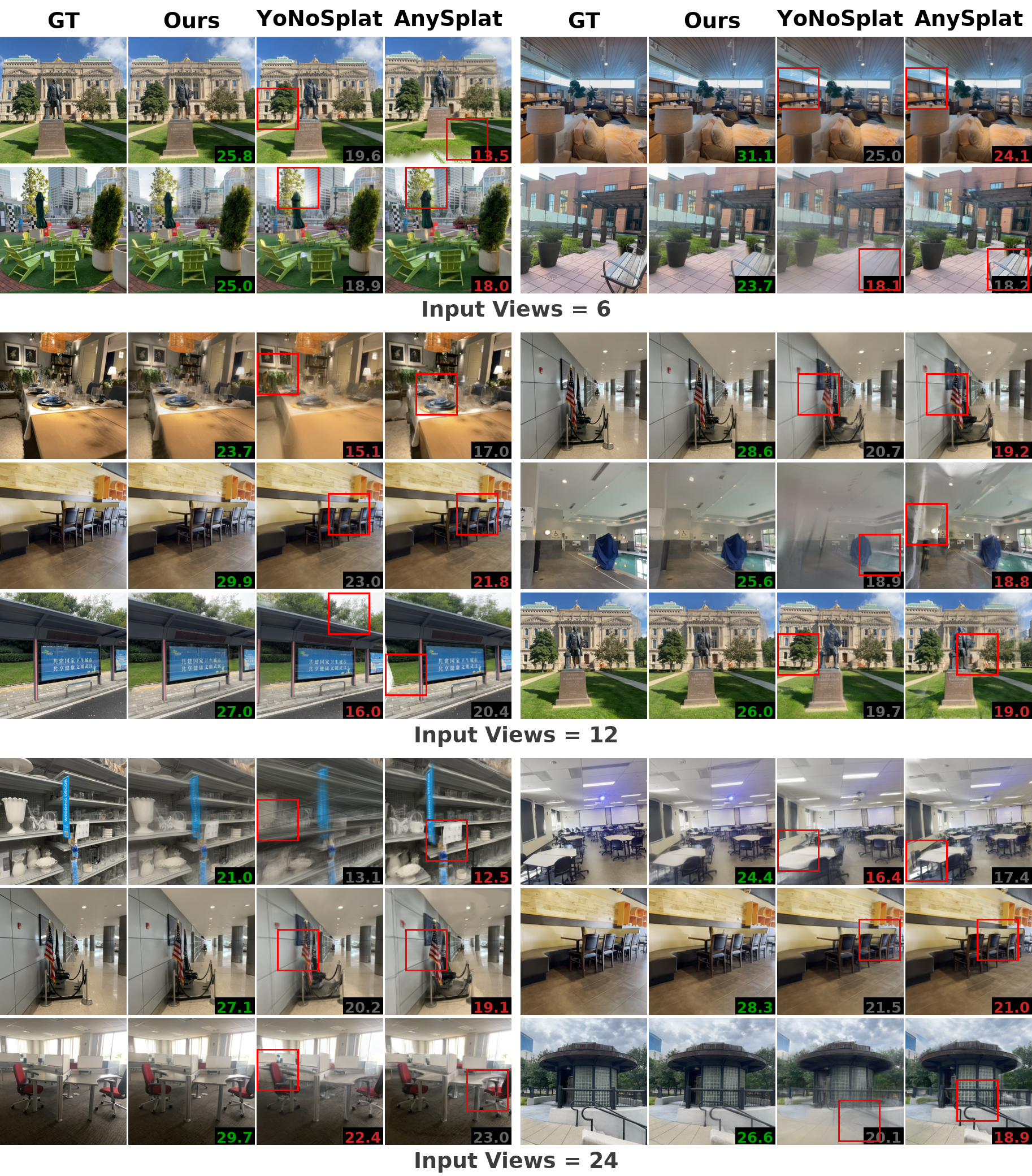}
  \caption{\textbf{Qualitative comparison of novel view synthesis on the DL3DV test set.} Here we report PSNR in the right corner in the pose-free, calibration-free setting. Our model produces higher-quality results compared to others. }
  \label{fig:dl3dv_visualization}

  
\end{figure}

\subsubsection{Baselines}
We compared against SOTA representative sparse-view generalizable methods on novel view synthesis: 1) Pose-dependent: MVSplat~\cite{chen2024mvsplat}, DepthSplat~\cite{xu2025depthsplat}; 2) Pose-free: NoPoSplat~\cite{ye2024noposplat}, AnySplat~\cite{jiang2025anysplat}, and YoNoSplat~\cite{ye2025yonosplat}. For relative pose estimation, we follow the evaluation method of Depth Anything v3~\cite{lin2025depthanythingv3} and then compare against SOTA methods: VGGT~\cite{wang2025vggt}, $\pi^3$~\cite{wang2025pi3}, Depth Anything v3 and YoNoSplat~\cite{ye2025yonosplat}.

\subsection{Pose Estimation and Novel-View Synthesis}
\subsubsection{Novel View Synthesis}

Table~\ref{tab:dl3dv_nvs} compares our method on the DL3DV dataset with representative feed-forward baselines for scene-level novel view synthesis, including pose-dependent methods (MVSplat and DepthSplat) and pose-free methods (NoPoSplat, AnySplat, and YoNoSplat). Our method consistently outperforms these baselines and improves over YoNoSplat by  1.6\,dB PSNR in the 24-view setting. In Table~\ref{tab:re10k_nvs}, the results show our method trained and evaluated on RE10K still surpasses the current SOTA method YoNoSplat.

Figure~\ref{fig:dl3dv_visualization} provides a qualitative visualization of DL3DV. Compared with YoNoSplat and AnySplat, our method produces sharper structures and cleaner textures under both short and long trajectories. In particular, the two baselines show more severe drift-induced degradation on long sequences, including blur, ghosting, and structural inconsistency, while our results remain visually stable with substantially reduced pose-drift artifacts.

\begin{table}[t]
  \centering
  \caption{Novel view synthesis comparison under various input settings on the DL3DV dataset. We report results with 6, 12, and 24 input views. Our model consistently achieves the best performance, where \textbf{$p$},\textbf{$k$} denote using ground-truth poses, intrinsics.}
  
  
  \label{tab:dl3dv_nvs}
  \begin{adjustbox}{width=0.9\textwidth, center}
  \begin{tabular}{l c | c c |ccc | ccc | ccc}
    \toprule
    &&&& \multicolumn{3}{c}{6v} & \multicolumn{3}{c}{12v} & \multicolumn{3}{c}{24v} \\
    \cmidrule(lr){5-7} \cmidrule(lr){8-10} \cmidrule(lr){11-13}
    Method & Venue & \textbf{$p$} & \textbf{$k$} & PSNR$\uparrow$ & SSIM$\uparrow$ & LPIPS$\downarrow$ & PSNR$\uparrow$ & SSIM$\uparrow$ & LPIPS$\downarrow$ & PSNR$\uparrow$ & SSIM$\uparrow$ & LPIPS$\downarrow$ \\
    \midrule
    MVSplat\cite{chen2024mvsplat} & ECCV'24 & \checkmark & \checkmark & 22.659 & 0.760 & 0.173 & 21.289 & 0.709 & 0.224 & 19.975 & 0.662 & 0.269 \\
    DepthSplat\cite{xu2025depthsplat} & CVPR'25 & \checkmark & \checkmark & 23.418 & 0.797 & 0.136 & 21.911 & 0.753 & 0.179 & 20.088 & 0.690 & 0.240 \\
    NoPoSplat\cite{ye2024noposplat} & ICLR'25 &  & \checkmark & 22.766 & 0.743 & 0.179 & 19.380 & 0.563 & 0.318 & 17.860 & 0.495 & 0.397 \\
    AnySplat\cite{jiang2025anysplat} & SA'25 (TOG) && & 19.027 & 0.554 & 0.235 & 18.940 & 0.549 & 0.262 & 19.703 & 0.596 & 0.249 \\
    YonoSplat\cite{ye2025yonosplat} & ICLR'26 && & 24.531 & 0.804 & 0.142 & 22.933 & 0.746 & 0.187 & 22.174 & 0.720 & 0.209 \\
    Erayzer$_{256\times256}$\cite{zhao2025erayzer} & CVPR'26 & & & 24.814 & 0.791 & 0.184 & 20.454 & 0.639 & 0.317 & 18.750 & 0.571 & 0.406 \\
    \rowcolor{gray!15}
    Ours & - && & \textbf{24.922} & \textbf{0.826} & \textbf{0.127} & \textbf{24.250} & \textbf{0.797} & \textbf{0.141} & \textbf{24.242} & \textbf{0.794} & \textbf{0.142} \\
    \bottomrule
  \end{tabular}
  \end{adjustbox}
\end{table}

\begin{figure}[t]
  \centering
  \includegraphics[width=\textwidth]{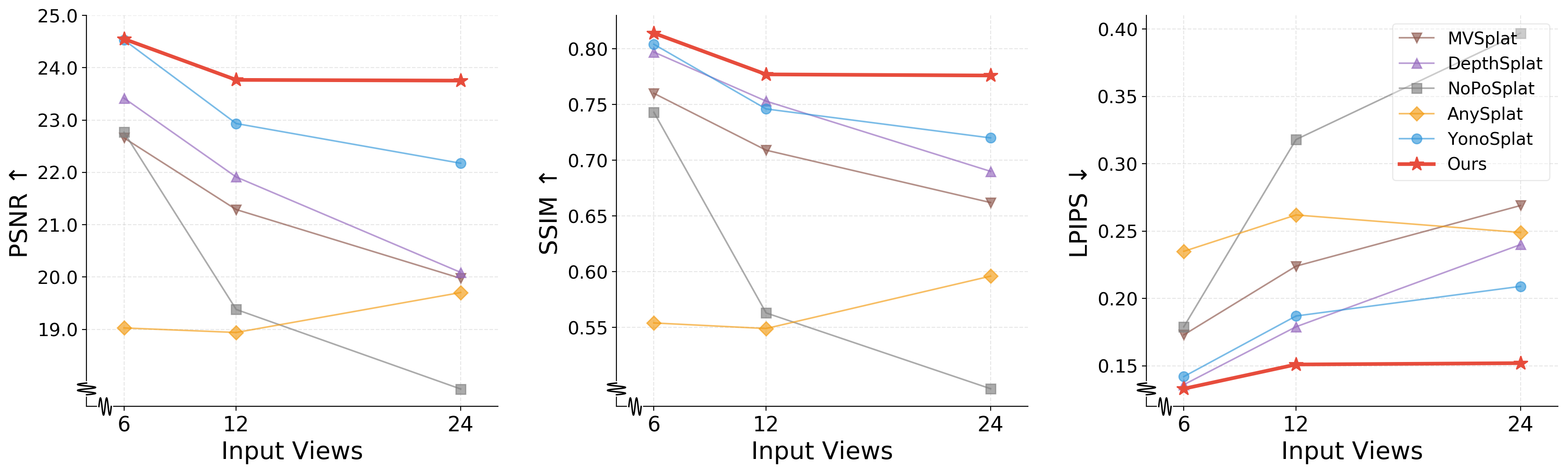}
  
  
  \caption{As the number of input views increases (6→12→24), YoNoSplat degrades markedly, while our method remains stable across all metrics, demonstrating stronger robustness to pose drift in long sequences.}
  \label{fig:metric}
  
\end{figure}

\begin{table}[t]
  \centering
  \caption{\textbf{Pose estimation comparison}. Our model achieves the best performance on the DL3DV dataset with an input resolution of 224$\times$224 under various input settings.}
  \label{tab:dl3dv_pose_auc}

  
  \resizebox{\linewidth}{!}{ 
  \begin{tabular}{l|ccc | ccc | ccc}
    \toprule
    & \multicolumn{3}{c}{6v} & \multicolumn{3}{c}{12v} & \multicolumn{3}{c}{24v} \\
    \cmidrule(lr){2-4} \cmidrule(lr){5-7} \cmidrule(lr){8-10}
    Model & AUC@5$^\circ$$\uparrow$ & AUC@10$^\circ$$\uparrow$ & AUC@20$^\circ$$\uparrow$ & AUC@5$^\circ$$\uparrow$ & AUC@10$^\circ$$\uparrow$ & AUC@20$^\circ$$\uparrow$ & AUC@5$^\circ$$\uparrow$ & AUC@10$^\circ$$\uparrow$ & AUC@20$^\circ$$\uparrow$ \\
    \midrule
    VGGT$_{518\times280}$~\cite{wang2025vggt} & 0.700 & 0.848 & 0.924 & - & - & - & - & - & - \\
    \(\pi^3\)$_{518\times280}$~\cite{wang2025pi3} & 0.795 & 0.897 & 0.949 & - & - & - & - & - & - \\
    NoPoSplat$_{256\times256}$~\cite{ye2024noposplat} & 0.538 & 0.735 & 0.853 & - & - & - & - & - & - \\
    AnySplat$_{448\times448}$~\cite{jiang2025anysplat} & 0.596 & 0.776 & 0.884 & 0.517 & 0.732 & 0.864 & 0.476 & 0.708 & 0.851 \\
    YonoSplat$_{224\times224}$~\cite{ye2025yonosplat} & 0.833 & 0.917 & 0.958 & 0.804 & 0.902 & 0.951 & 0.778 & 0.885 & 0.942 \\
    Erayzer$_{256\times256}$\cite{zhao2025erayzer} & 0.846 & 0.972 & 0.983 & 0.421 & 0.699 & 0.767 & 0.389 & 0.654 & 0.707 \\
    \rowcolor{gray!15}
    Ours$_{224\times224}$ & \textbf{0.967} & \textbf{0.983} & \textbf{0.992} & \textbf{0.961} & \textbf{0.979} & \textbf{0.990} & \textbf{0.949} & \textbf{0.972} & \textbf{0.985} \\
    \bottomrule
  \end{tabular}
  }
  
\end{table}


\begin{table}[t]
  \begin{minipage}{0.48\textwidth}
  \centering
  \caption{\textbf{Zero-shot pose estimation on the RE10k dataset (6 input views)}. We trained the model on DL3DV and tested on RE10k dataset.}

  
  \label{tab:re10k_pose_auc}
  \resizebox{\textwidth}{!}{
  \begin{tabular}{lccc}
    \toprule
    Method & AUC@5$^\circ$$\uparrow$ & AUC@10$^\circ$$\uparrow$ & AUC@20$^\circ$$\uparrow$ \\
    \midrule
    MASt3R$_{518\times288}$ & 0.609 & 0.776 & 0.878 \\
    VGGT$_{518\times280}$ & 0.566 & 0.753 & 0.867 \\
    \(\pi^3\)$_{518\times280}$ & 0.705 & 0.841 & 0.916 \\
    NoPoSplat$_{256\times256}$ & 0.443 & 0.627 & 0.755 \\
    YoNoSplat$_{224\times224}$ & 0.740 & 0.895 & \textbf{0.924} \\
    \rowcolor{gray!15}
    Ours$_{224\times224}$ & \textbf{0.755} & \textbf{0.911} & 0.923 \\
    \bottomrule
  \end{tabular}
  }
  \end{minipage}
  \hfill
  \begin{minipage}{0.48\textwidth}
  \centering
  \caption{\textbf{NVS comparison on the RE10k dataset (6 input views).} We trained the model on the RE10k dataset and tested on it.}

  
  \label{tab:re10k_nvs}
  \resizebox{0.8\textwidth}{!}{
  \begin{tabular}{lcc|ccc}
    \toprule
    Method & \textbf{$p$} & \textbf{$k$} & PSNR$\uparrow$ & SSIM$\uparrow$ & LPIPS$\downarrow$ \\
    \midrule
    DepthSplat & \checkmark & \checkmark & 24.156 & 0.846 & 0.145 \\
    NoPoSplat  &           & \checkmark & 22.175 & 0.750 & 0.207 \\
    YoNoSplat       & \checkmark & \checkmark & 25.037 & 0.848 & 0.134 \\
    YoNoSplat       &           & \checkmark & 25.395 & 0.857 & 0.131 \\
    YoNoSplat       &           &           & 24.571 & 0.823 & 0.144 \\
    \rowcolor{gray!15}
    Ours &&& \textbf{25.736} & \textbf{0.876} & \textbf{0.128} \\
    \bottomrule
  \end{tabular}
  }
  \end{minipage}
\end{table}



\subsubsection{Camera Pose Estimation}

Table~\ref{tab:dl3dv_pose_auc} reports a clear and consistent advantage of our method in camera pose estimation across all view settings. We attribute this gain to our raymap-guided representation, which injects explicit geometric constraints into training and effectively \emph{disambiguates the coupled optimization of geometry and pose}, leading to state-of-the-art pose accuracy. Notably, the improvement is more pronounced as the number of input views increases (12v/24v), where trajectories are longer, and drift accumulation is more severe; this trend verifies that our method better \emph{controls cumulative pose drift} and maintains globally stable camera estimation in long-sequence scenarios.

\subsubsection{Cross-Dataset Generalization}

To evaluate cross-dataset generalization, we train the model on DL3DV and directly evaluate on ScanNet++ without any fine-tuning. We compare against AnySplat and YoNoSplat on rendering performance. Notably, our method remains favorable even despite AnySplat's training-domain advantage. As shown in Table~\ref{tab:scannetpp_generalization}, our approach consistently achieves state-of-the-art performance across all input-view settings.

We further evaluate zero-shot pose generalization in Table~\ref{tab:re10k_pose_auc} by training on DL3DV and testing on RE10K. Without target-domain fine-tuning, our method achieves the best on 5$^\circ$ and 10$^\circ$, demonstrating cross-dataset transferability of the proposed raymap-guided representation for pose estimation.

\begin{figure}[t]
  \centering
  \includegraphics[width=\textwidth]{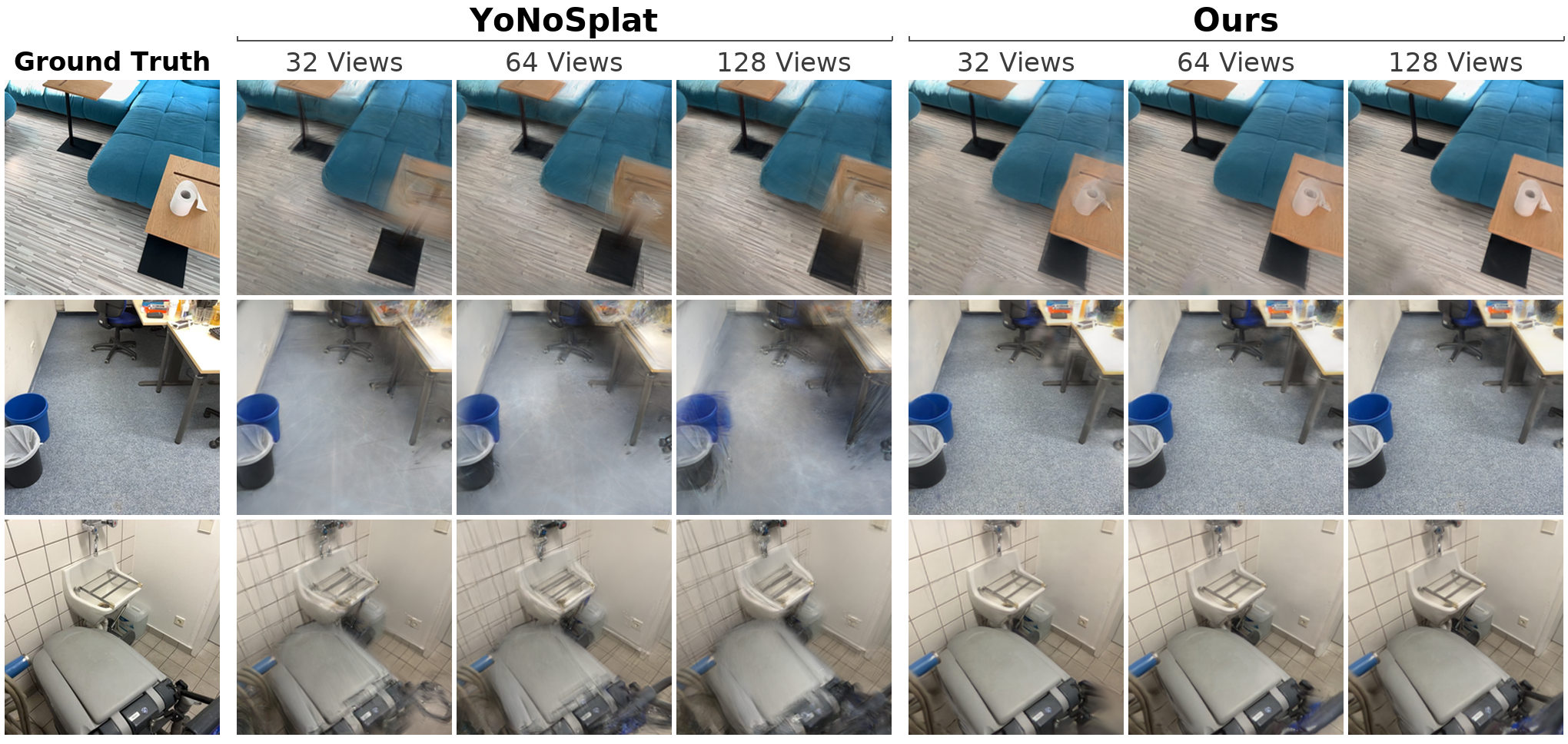}

  
  \caption{Our model generalizes better than YoNoSplat on the ScanNet++ dataset and demonstrates coherent fusion of Gaussians across longer sequences. We attribute this gain to our synergistic framework, which tightly couples geometry and appearance.}
  \label{fig:dl3dv_visualization}

  
\end{figure}

\begin{table}[t]
  \centering
  \caption{\textbf{Generalization to ScanNet++.} Trained on DL3DV and tested on ScanNet++, our model outperforms YonoSplat~\cite{ye2025yonosplat}. Input images are sampled from full sequences with a fixed target view. Our performance improves more on long-sequences because of accurate pose estimation.}

  
  \label{tab:scannetpp_generalization}
  \begin{adjustbox}{width=0.8\textwidth, center}
  \begin{tabular}{lccccccccc}
    \toprule
    & \multicolumn{3}{c}{32v}
    & \multicolumn{3}{c}{64v}
    & \multicolumn{3}{c}{128v} \\
    \cmidrule(lr){2-4} \cmidrule(lr){5-7} \cmidrule(lr){8-10}
    Method & PSNR$\uparrow$ & SSIM$\uparrow$ & LPIPS$\downarrow$
    & PSNR$\uparrow$ & SSIM$\uparrow$ & LPIPS$\downarrow$
    & PSNR$\uparrow$ & SSIM$\uparrow$ & LPIPS$\downarrow$ \\
    \midrule
    AnySplat       & 14.054 & 0.494 & 0.468 & 15.982 & 0.551 & 0.412 & 16.988 & 0.583 & 0.386 \\
    YoNoSplat & 16.886 & 0.600 & 0.432 & 17.368 & 0.608 & 0.413 & 17.641 & 0.617 & 0.405 \\
    \rowcolor{gray!15}
    Ours           & \textbf{17.569} & \textbf{0.620} & \textbf{0.419}
                   & \textbf{18.935} & \textbf{0.668} & \textbf{0.348}
                   & \textbf{19.714} & \textbf{0.695} & \textbf{0.317} \\
    \bottomrule
  \end{tabular}
  \end{adjustbox}

  
\end{table}

\subsection{Ablation Studies}

\subsubsection{Ablations on Raymap-Guided Coupling Module(RGC)}

In ~\cref{tab:loss_ablation_all}, the first row (w/o raymap-guided) replaces raymap-guided geometry with geometry derived from camera-head pose prediction. Although this variant can still produce competitive rendering in some settings, its pose accuracy drops consistently versus the full model, showing that raymap-guided geometry is crucial for accurate pose estimation and global geometric consistency.

The second and third rows maintain the geometry guided by the raymap. Removing raymap loss (w/o raymap loss) weakens rendering quality, indicating that stronger geometry constraints from raymaps improve appearance (geometry helps appearance). Conversely, removing rendering loss (w/o rendering loss) degrades pose accuracy, showing that appearance supervision is also necessary to stabilize and refine geometry/pose (appearance helps geometry). Together, these trends verify the bidirectional coupling in our framework: geometry supervision improves rendering, and rendering supervision improves geometry. This mutual reinforcement is the \emph{central insight} of our method and the main reason why our full model is consistently robust across view settings.

\begin{table}[t]
  \centering
  \caption{Loss ablation across different input views on the large model.}

  
  \label{tab:loss_ablation_all}
  \resizebox{\linewidth}{!}{%
  \begin{tabular}{l|cccc | cccc | cccc}
    \toprule
    & \multicolumn{4}{c}{6v} & \multicolumn{4}{c}{12v} & \multicolumn{4}{c}{24v} \\
    \cmidrule(lr){2-5} \cmidrule(lr){6-9} \cmidrule(lr){10-13}
    Model & PSNR$\uparrow$ & SSIM$\uparrow$ & AUC@5$^\circ$$\uparrow$ & AUC@10$^\circ$$\uparrow$ &
    PSNR$\uparrow$ & SSIM$\uparrow$ & AUC@5$^\circ$$\uparrow$ & AUC@10$^\circ$$\uparrow$ &
    PSNR$\uparrow$ & SSIM$\uparrow$ & AUC@5$^\circ$$\uparrow$ & AUC@10$^\circ$$\uparrow$ \\
    \midrule
    w/o raymap-guided & \underline{23.062} & \underline{0.752} & 0.821 & 0.907 & \textbf{22.375} & \textbf{0.717} & 0.753 & 0.867 & \underline{21.379} & \underline{0.599} & 0.690 & 0.821 \\
    w/o raymap loss & 21.869 & 0.708 & \underline{0.869} & \underline{0.927} & 20.933 & 0.655 & \underline{0.814} & \underline{0.902} & 20.609 & 0.635 & \underline{0.782} & \underline{0.875} \\
    w/o rgb loss & 12.608 & 0.321 & 0.851 & 0.922 & 12.375 & 0.325 & 0.801 & 0.891 & 12.378 & 0.328 & 0.754 & 0.851 \\
    Ours & \textbf{23.302} & \textbf{0.766} & \textbf{0.874} & \textbf{0.935} & \underline{21.909} & \underline{0.696} & \textbf{0.829} & \textbf{0.907} & \textbf{21.626} & \textbf{0.675} & \textbf{0.787} & \textbf{0.883} \\
    \bottomrule
  \end{tabular}%
  }
\end{table}

\subsubsection{Dual-Frequency Viewpoint Scheduling Sampler}

Table~\ref{tab:sampler_ablation} provides three findings. (1) Our sampler consistently outperforms the original sampler (similar to YoNoSplat), indicating sampling strategy is a primary limiting factor. (2) Without replay, increasing $g_{\max}$ yields a clear trade-off: performance degrades on the short range (0--50) but improves on the long range (0--150). This suggests that a a larger interval expansion improves long-range coverage at the cost of short-range consistency. (3) Adding replay at $g_{\max}=\infty$ preserves strong long-range performance while recovering short-range performance. Overall, replay mitigates the short/long-range trade-off and enables a better balance between short-range and long-range fidelity.

\begin{table}[t]
    \centering
    \caption{Ablation on \emph{replay} and $g_{\max}$ settings.}

    
    \label{tab:sampler_ablation}
    \resizebox{\linewidth}{!}{
        \begin{tabular}{l|cccc|cccc}
        \toprule
        & \multicolumn{4}{c}{6 view, range 0--50} & \multicolumn{4}{c}{6 view, range 0--150} \\
        \cmidrule(lr){2-5} \cmidrule(lr){6-9}
        Model & PSNR$\uparrow$ & SSIM$\uparrow$ & AUC@5$^\circ$$\uparrow$ & AUC@10$^\circ$$\uparrow$ & PSNR$\uparrow$ & SSIM$\uparrow$ & AUC@5$^\circ$$\uparrow$ & AUC@10$^\circ$$\uparrow$ \\
        \midrule
        original sampler & 22.696 & 0.747 & 0.782 & 0.884 & 18.815 & 0.571 & 0.591 & 0.744 \\
        $g_{\max}=10$, w/o \emph{replay} & 23.302 & 0.766 & 0.874 & 0.935 & 19.542 & 0.579 & 0.687 & 0.803 \\
        $g_{\max}=20$, w/o \emph{replay} & 22.892 & 0.752 & 0.872 & 0.935 & 19.690 & 0.592 & 0.709 & 0.826 \\
        $g_{\max}=\infty$, w/o \emph{replay} & 22.744 & 0.742 & 0.862 & 0.930 & 19.712 & 0.588 & 0.722 & 0.832 \\
        $g_{\max}=\infty$, $\emph{replay}=50\%$ & 23.006 & 0.745 & 0.864 & 0.929 & 19.745 & 0.592 & 0.721 & 0.832 \\
        \bottomrule
        \end{tabular}
    }
\end{table}

\section{Conclusion}
In this paper, we introduced a synergistic framework with a Raymap-Guided Coupling Module (RGC), which tightly couples geometric and appearance supervision. To stabilize training across wide temporal intervals, we further proposed a Dual-Frequency Viewpoint Scheduling strategy that combines easy-to-hard interval expansion with replay of short-interval samples. Extensive results and ablations show that the proposed bidirectional geometry–appearance coupling is the key to achieving robust, scalable pose-free 3D reconstruction with stronger rendering and pose performance on long-sequence generalization.
\subsubsection{Limitations and Discussions.}
In the appendix, we use the same hyperparameters across all datasets for multi-dataset training. Although this  generalizes well, an adaptive mechanism that can automatically adjust the sampling strategy across datasets would be a future direction, as it could improve performance and reduce tuning costs. Furthermore, our pipeline is pose-free at inference, but still relies on pose and raymap supervision during training. Extending it to a unified framework that can handle posed and unposed training data with less supervision is also another important direction for future work.


\section*{Acknowledgements}

This research was supported by the Ministry of Science and ICT (MSIT) of Korea, under the National Research Foundation (NRF) grant (RS-2024-00337548); supported by a grant of the Korea-US Collaborative Research Fund (KUCRF), funded by the Ministry of Science and ICT and Ministry of Health \& Welfare, Republic of Korea (grant number: RS-2024-00468417). This work was also supported by the Institute of Information \& communications Technology Planning \& Evaluation (IITP) grant funded by the Korean government (MSIT) (No. RS-2024-00457882, AI Research Hub Project) and the Institute for Information \& Communications Technology Planning \& Evaluation (IITP) grant funded by the Korea government(MSIT)  (No.RS-2025-25441838, Development of a human foundation model for human-centric universal artificial intelligence and training of personnel). This research was also supported by the Advanced GPU Utilization Support Program funded by the Government of the Republic of Korea (Ministry of Science and ICT) and the Institute of Information \& Communications Technology Planning \& Evaluation(IITP) grant funded by the Korea government(MSIT) (No.RS-2026-25517417, Development of Compression, Reconstruction, and Rendering Technologies for Free-viewpoint Media).



%
%
\bibliographystyle{splncs04}
\bibliography{main}










\clearpage

\title{Raymap-Guided Coupling for Drift-Robust Unposed Feed-Forward 3D Reconstruction} 

\titlerunning{Raymap-Guided Coupling}


\authorrunning{X. Sun et al.}


\thispagestyle{empty}
\appendix
\begin{figure}[H]
  \centering
  \includegraphics[width=\textwidth]{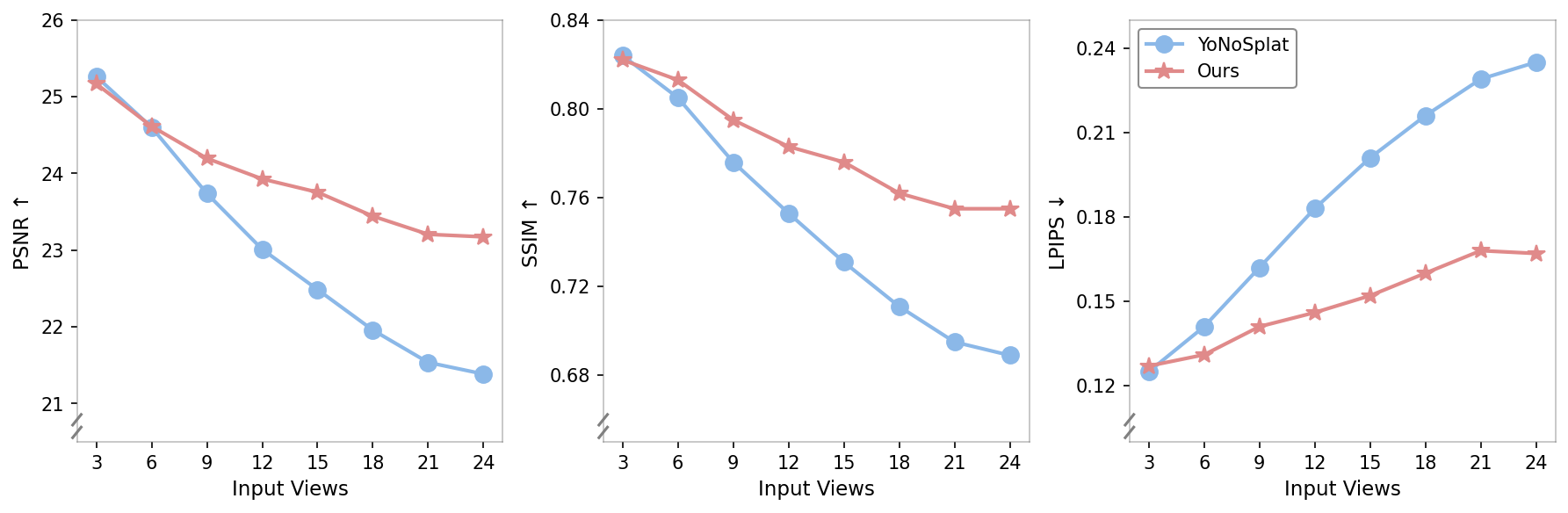}
  \caption{\textbf{Pose drift impact on novel view synthesis under long input sequences.} As the number of input views increases, cumulative pose estimation errors significantly degrade the rendering quality of YoNoSplat. In contrast, our method maintains stable performance across longer sequences, demonstrating strong robustness to pose drift.}
  \label{fig:supple_pose_drift}
\end{figure}

\section{Appendix}

\noindent {\bf Pose drift becomes increasingly problematic when reconstructing scenes from long image sequences.}

To evaluate robustness under this setting, we perform inference with progressively increasing numbers of input views. As shown in Fig. 1, the performance of YoNoSplat~\cite{ye2025yonosplat} deteriorates significantly as the sequence length increases, indicating severe accumulation of pose errors. In contrast, our method maintains substantially more stable performance across longer sequences, demonstrating its ability to effectively suppress pose drift and preserve rendering quality for novel view synthesis.


\noindent {\bf Pose Estimation Comparison with Finetuning Depth Anything v3.}

To fairly compare our performance with DepthAnything v3~\cite{lin2025depthanythingv3}, we report pose estimation results using DepthAnything v3 giant model, as well as results obtained after finetuning it on the DL3DV~\cite{ling2024dl3dv} dataset in ~\cref{tab:supple_pose_auc}. Our new model consistently beat finetuned DepthAnything v3 based on the synergistic framework.


\begin{table}[H]
  \centering
  \caption{Pose estimation comparison. $*$ means finetune DepthAnything v3 on the DL3DV dataset.}
  \label{tab:supple_pose_auc}
  \resizebox{\linewidth}{!}{
  \begin{tabular}{l|ccc | ccc | ccc}
    \toprule
    & \multicolumn{3}{c}{6v} & \multicolumn{3}{c}{12v} & \multicolumn{3}{c}{24v} \\
    \cmidrule(lr){2-4} \cmidrule(lr){5-7} \cmidrule(lr){8-10}
    Model & AUC@5$^\circ$$\uparrow$ & AUC@10$^\circ$$\uparrow$ & AUC@20$^\circ$$\uparrow$ & AUC@5$^\circ$$\uparrow$ & AUC@10$^\circ$$\uparrow$ & AUC@20$^\circ$$\uparrow$ & AUC@5$^\circ$$\uparrow$ & AUC@10$^\circ$$\uparrow$ & AUC@20$^\circ$$\uparrow$ \\
    \midrule
    NoPoSplat$_{256\times256}$~\cite{ye2024noposplat} & 0.538 & 0.735 & 0.853 & - & - & - & - & - & - \\
    AnySplat$_{448\times448}$~\cite{jiang2025anysplat} & 0.596 & 0.776 & 0.884 & 0.517 & 0.732 & 0.864 & 0.476 & 0.708 & 0.851 \\
    YonoSplat$_{224\times224}$~\cite{ye2025yonosplat} & 0.833 & 0.917 & 0.958 & 0.804 & 0.902 & 0.951 & 0.778 & 0.885 & 0.942 \\
    \midrule
    DepthAnything v3$_{224\times224}$ & 0.659 & 0.823 & 0.911 & 0.576 & 0.769 & 0.879 & 0.529 & 0.736 & 0.860 \\
    \rowcolor{gray!15}
    DepthAnything v3$*$$_{224\times224}$ & 0.955 & 0.978 & 0.989 & 0.928 & 0.953 & 0.980 & 0.920 & 0.956 & 0.977\\
    \rowcolor{gray!15}
    Ours$_{224\times224}$ & \textbf{0.970} & \textbf{0.985} & \textbf{0.992} & \textbf{0.942} & \textbf{0.969} & \textbf{0.984} & \textbf{0.936} & \textbf{0.966} & \textbf{0.982} \\
    \bottomrule
  \end{tabular}
  }
\end{table}

\noindent {\bf More Details on Loss Function.}
The synergistic loss objective is a weighted sum as follows:
\begin{equation}
  \mathcal{L} = \mathcal{L}_{\text{rgb}} + \lambda_{\text{cam}} \mathcal{L}_{\text{cam}} + \lambda_{\text{ray}} \mathcal{L}_{\text{ray}}.
\end{equation}
where $\lambda_{\text{ray}}=1.0$ and $\lambda_{\text{cam}}=0.5$.

Furthermore, for $\mathcal{L}_{\text{ray}}$, let $\Omega=\{1,\dots,H\}\times\{1,\dots,W\}$ be the image grids, and let
$\mathbf{x}_{u,v}\in\mathbb{R}^2$ denote the normalized pixel coordinates.
For each context view index $m$, we compute world-space rays from camera extrinsics $\mathbf{E}_{m}$ and intrinsics $\mathbf{K}_{m}$:

\begin{equation}
(\mathbf{o}_{m,u,v}, \mathbf{d}_{m,u,v})
= \mathrm{Ray}(\mathbf{x}_{u,v}; \mathbf{K}_m, \mathbf{E}_m),
\quad (u,v)\in\Omega,
\end{equation}

where $\mathrm{Ray}(\cdot)$ computes the origin $\mathbf{o}_{m,u,v}\in\mathbb{R}^3$ and direction $\mathbf{d}_{m,u,v}\in\mathbb{R}^3$ of the camera ray 
in the world-space
corresponding to the pixel coordinate $\mathbf{x}_{u,v}$ given the 
intrinsic matrix of the camera $\mathbf{K}_m$ and the extrinsic matrix $\mathbf{E}_m$.

Given the predicted DA3 ray tensor
$\hat{\mathbf{r}}_{m,u,v}\in\mathbb{R}^6$, we split it into
\begin{equation}
\hat{\mathbf{d}}_{m,u,v}=\hat{\mathbf{r}}_{m,u,v,0:3},
\qquad
\hat{\mathbf{o}}_{m,u,v}=\hat{\mathbf{r}}_{m,u,v,3:6}.
\end{equation}
We supervise origins and directions with MSE losses:
\begin{equation}
\mathcal{L}_{\mathrm{origin}}
=
\frac{1}{MHW}\sum_{m=1}^{M}\sum_{(u,v)\in\Omega}
\left\|
\mathbf{o}_{m,u,v}-\hat{\mathbf{o}}_{m,u,v}
\right\|_2^2,
\end{equation}
\begin{equation}
\mathcal{L}_{\mathrm{direction}}
=
\frac{1}{MHW}\sum_{m=1}^{M}\sum_{(u,v)\in\Omega}
\left\|
\mathbf{d}_{m,u,v}-\hat{\mathbf{d}}_{m,u,v}
\right\|_2^2.
\end{equation}
Optionally, the ray loss is
\begin{equation}
\mathcal{L}_{\mathrm{ray}}
= \mathcal{L}_{\mathrm{origin}} + \mathcal{L}_{\mathrm{direction}}.
\end{equation}

\begin{table}
    \centering
    \captionof{table}{Novel view synthesis and pose estimation comparison on the DL3DV dataset.}
    \begin{adjustbox}{max width=\linewidth}
    \setlength{\tabcolsep}{0.08cm}
    \renewcommand{\arraystretch}{1.0}
    
    \begin{tabular}{l c | cccc | cccc | cccc }
        \toprule
        && \multicolumn{4}{c}{6 views} & \multicolumn{4}{c}{12 views} & \multicolumn{4}{c}{24 views} \\
        \cmidrule(lr){3-6} \cmidrule(lr){7-10} \cmidrule(lr){11-14}
        Method & Params & speed & PSNR$\uparrow$ & AUC@5$^\circ$$\uparrow$ & AUC@10$^\circ$$\uparrow$ &  speed & PSNR$\uparrow$ & AUC@5$^\circ$$\uparrow$ & AUC@10$^\circ$$\uparrow$ &  speed & PSNR$\uparrow$ & AUC@5$^\circ$$\uparrow$ & AUC@10$^\circ$$\uparrow$ \\
        \midrule
        AnySplat$_{448\times448}$ & 1.19B & 0.955s & 19.027 & 0.596 & 0.776 & 2.101s & 18.940 & 0.517 & 0.732 & 3.940s & 19.703 & 0.476 & 0.708 \\
        YonoSplat$_{224\times224}$ & 1.02B & 0.236s & \underline{24.531} & 0.833 & 0.917 & 0.317s & \underline{22.933} & 0.804 & 0.902 & 0.395s & \underline{22.174} & 0.778 & 0.885 \\
        \midrule
        \textbf{Ours-Large} & 0.44B & 0.108s & 24.341 & \underline{0.906} & \underline{0.950} & 0.172s & 22.729 & \underline{0.841} & \underline{0.913} & 0.319s & 22.095 & \underline{0.788} & \underline{0.887} \\
        \textbf{Ours-Giant} & 1.40B & 0.155s & \textbf{24.550} & \textbf{0.970} & \textbf{0.985} & 0.297s & \textbf{23.767} & \textbf{0.942} & \textbf{0.969} & 0.667s & \textbf{23.753} & \textbf{0.936} & \textbf{0.966} \\
        \bottomrule
    \end{tabular}
    \end{adjustbox}
\label{tab:runtime_psnr}
\end{table}

\begin{table}[!t]
    \centering
    \caption{Pose estimation comparison with 3d foundation models.}
    \begin{adjustbox}{max width=\linewidth}
    \setlength{\tabcolsep}{0.08cm}
    \renewcommand{\arraystretch}{1.0}
    \begin{tabular}{l|ccc | ccc | ccc}
        \toprule
        & \multicolumn{3}{c}{6 views} & \multicolumn{3}{c}{12 views} & \multicolumn{3}{c}{24 views} \\
        \midrule
        Method & 5$^\circ$$\uparrow$ & 10$^\circ$$\uparrow$ & 20$^\circ$$\uparrow$ & 5$^\circ$$\uparrow$ & 10$^\circ$$\uparrow$ & 20$^\circ$$\uparrow$ & 5$^\circ$$\uparrow$ & 10$^\circ$$\uparrow$ & 20$^\circ$$\uparrow$ \\
        \midrule
        VGGT & 0.700 & 0.848 & 0.924 & - & - & - & - & - & - \\
        \(\pi^3\) & 0.795 & 0.897 & 0.949 & - & - & - & - & - & -  \\
        \rowcolor{gray!20}
        DAv3 & 0.659 & 0.823 & 0.911 & 0.576 & 0.769 & 0.879 & 0.529 & 0.736 & 0.860 \\
        \textbf{Ours} & \textbf{0.970} & \textbf{0.985} & \textbf{0.992} & \textbf{0.942} & \textbf{0.969} & \textbf{0.984} & \textbf{0.936} & \textbf{0.966} & \textbf{0.982} \\
        \bottomrule
    \end{tabular}
    \end{adjustbox}
\label{tab:dav3_base}
\end{table}

\begin{table}[!t]
    \centering
    \caption{Ablation studies on 6 views based on our RGC Module.}
    \begin{adjustbox}{max width=\linewidth}
    \setlength{\tabcolsep}{0.08cm}
    \renewcommand{\arraystretch}{1.0}
    \begin{tabular}{l|cccccc}
        \toprule
        View Nums (DL3DV dataset) & PSNR$\uparrow$ & SSIM$\uparrow$ & LPIPS$\downarrow$ & 5$^\circ$$\uparrow$ & 10$^\circ$$\uparrow$ & 20$^\circ$$\uparrow$ \\
        \midrule
        \rowcolor{gray!20}
        DAv3 Gaussian Head & 22.407 & 0.731 & 0.212 & 0.841 & 0.919 & 0.959 \\
        Ours detach raymap and depth & \multicolumn{6}{c}{does not converge} \\
        Ours detach raymap & 20.178 & 0.610 & 0.320 & 0.833 & 0.914 & 0.957 \\
        \textbf{Ours} & \textbf{23.302} & \textbf{0.766} & \textbf{0.168} & \textbf{0.874} & \textbf{0.935} & \textbf{0.967} \\
        \bottomrule
    \end{tabular}
    \end{adjustbox}
\label{tab:GSHead_ablation}
\end{table}

\begin{table}[t]
  \centering
  \caption{Loss ablation on the large model for the DL3DV dataset.}
  
  \label{tab:loss_ablation_small}
  \resizebox{\linewidth}{!}{%
  \begin{tabular}{l|cccc | cccc }
    \toprule
    & \multicolumn{4}{c}{6views} & \multicolumn{4}{c}{12views} \\
    \cmidrule(lr){2-5} \cmidrule(lr){6-9}
    Model & PSNR$\uparrow$ & SSIM$\uparrow$ & AUC@5$^\circ$$\uparrow$ & AUC@10$^\circ$$\uparrow$ &
    PSNR$\uparrow$ & SSIM$\uparrow$ & AUC@5$^\circ$$\uparrow$ & AUC@10$^\circ$$\uparrow$ \\
    \midrule
    w/o raymap loss & 21.869 & 0.708 & 0.869 & 0.927 & 20.933 & 0.655 & 0.814 & 0.902  \\
    \rowcolor{gray!20}
    w/o rgb loss & 12.608 & 0.321 & 0.851 & 0.922 & 12.375 & 0.325 & 0.801 & 0.891 \\
    Ours & 23.302 & 0.766 & 0.874 & 0.935 & 21.909 & 0.696 & 0.829 & 0.907 \\
    \bottomrule
  \end{tabular}%
  }
  
\end{table}

\begin{table}[!t]
    \centering
    \caption{Zero-shot generalization on the Mip-NeRF360 dataset.}
    \begin{adjustbox}{max width=\linewidth}
    \setlength{\tabcolsep}{0.08cm}
    \renewcommand{\arraystretch}{1.0}
    \begin{tabular}{lc | cccc | cccc}
        \toprule
        && \multicolumn{4}{c}{Indoor Scene} & \multicolumn{4}{c}{Ourdoor Scene} \\
        \cmidrule(lr){3-6} \cmidrule(lr){7-10}
        Method & Params  & PSNR$\uparrow$ & SSIM$\uparrow$ & AUC@10$^\circ$$\uparrow$ & AUC@20$^\circ$$\uparrow$ & PSNR$\uparrow$ & SSIM$\uparrow$ & AUC@10$^\circ$$\uparrow$ & AUC@20$^\circ$$\uparrow$ \\
        \midrule
        AnySplat & 1.19B  & 19.76 & 0.518 & 0.667 & 0.815 & 13.50 & 0.231 & 0.669 & 0.782 \\
        E-Rayzer & 0.25B & 20.05 & 0.574 & 0.464 & 0.653 & 16.19 & 0.248 & 0.106 & 0.525 \\ 
        \textbf{Ours-Large} & 0.44B & 20.54 & 0.590 & 0.842 & 0.921 & 17.52 & 0.288 & 0.767 & 0.883 \\
        \bottomrule
    \end{tabular}
    \end{adjustbox}
\label{tab:zero_shot}
\end{table}

\noindent\textbf{Fair comparison of inference speed.} We report end-to-end inference speed and model size in ~\cref{tab:runtime_psnr} for fair comparison. Inference speed is measured end-to-end on one A100 GPU. Ours-Giant consistently achieves the best rendering quality and pose accuracy. Meanwhile, Ours-Large offers the fastest inference speed while maintaining strong performance, highlighting an efficient quality–speed tradeoff. 

\noindent\textbf{Clarification on pose-free inference vs. training.}
We clearly stated here that our method is pose-free at inference, but not pose-free or self-supervised during training, since it uses pose and raymap supervision with COLMAP-based camera poses. Our method does not remove pose estimation, but makes feed-forward reconstruction more robust under pose uncertainty through stronger supervision and a better geometric representation.

\noindent\textbf{Raymap-guided Coupling for 3DGS.} Depth Anything v3 (DAv3) introduces a geometric predictor based on raymap and depth, its performance remains relatively limited in ~\cref{tab:dav3_base}, suggesting that our gains are not merely inherited from a strong baseline. Our key contribution is replacing coupling through low-dimensional pose with a denser geometric interface via raymaps. 
The key improvement comes from our Raymap-Guided Coupling mechanism, which couples raymap and depth to reconstruct 3D Gaussians and enables more effective joint optimization of geometry, rendering, and pose. 

In ~\cref{tab:GSHead_ablation}, replacing our coupling design with the DAv3 gaussian head significantly degrades both rendering quality and pose accuracy, even though both variants rely on the same pre-trained 3D prior. Moreover, detaching the raymap branch in our gaussian head causes a clear performance drop; and detaching both raymap and depth gradients collapses training. More broadly, our evidence shows that coupling on raymap and depth is a stronger representation for 3DGS than pointmap-only representations (Uni3R) or alternatives based on depth and pose coupling (DAv3). These results indicate that the benefit comes from our coupling mechanism rather than pretrained initialization. 

Furthermore, the loss ablations in~\cref{tab:loss_ablation_small} show that joint rendering and geometry supervision is also important. Compared with the variant without rgb loss, i.e., using only geometric supervision, our full objective yields substantially better rendering quality and pose accuracy. This further supports that the gain comes from our full 3DGS reconstruction and rendering-based optimization.

\noindent\textbf{Computational Overhead.}
Our training is time-consuming, inference runs within seconds on a RTX 5090, indicating a much lower deployment cost.

\noindent\textbf{Evaluation Results on the multi-dataset training.}
We additionally report an Ours-Large model trained on the same seven mixed datasets as E-Rayzer. As shown in~\cref{tab:zero_shot}, this model outperforms E-Rayzer and AnySplat on the Mip-NeRF360 dataset, both indoor and outdoor scenes. 

\begin{figure}[H]
  \centering
  \includegraphics[width=\textwidth]{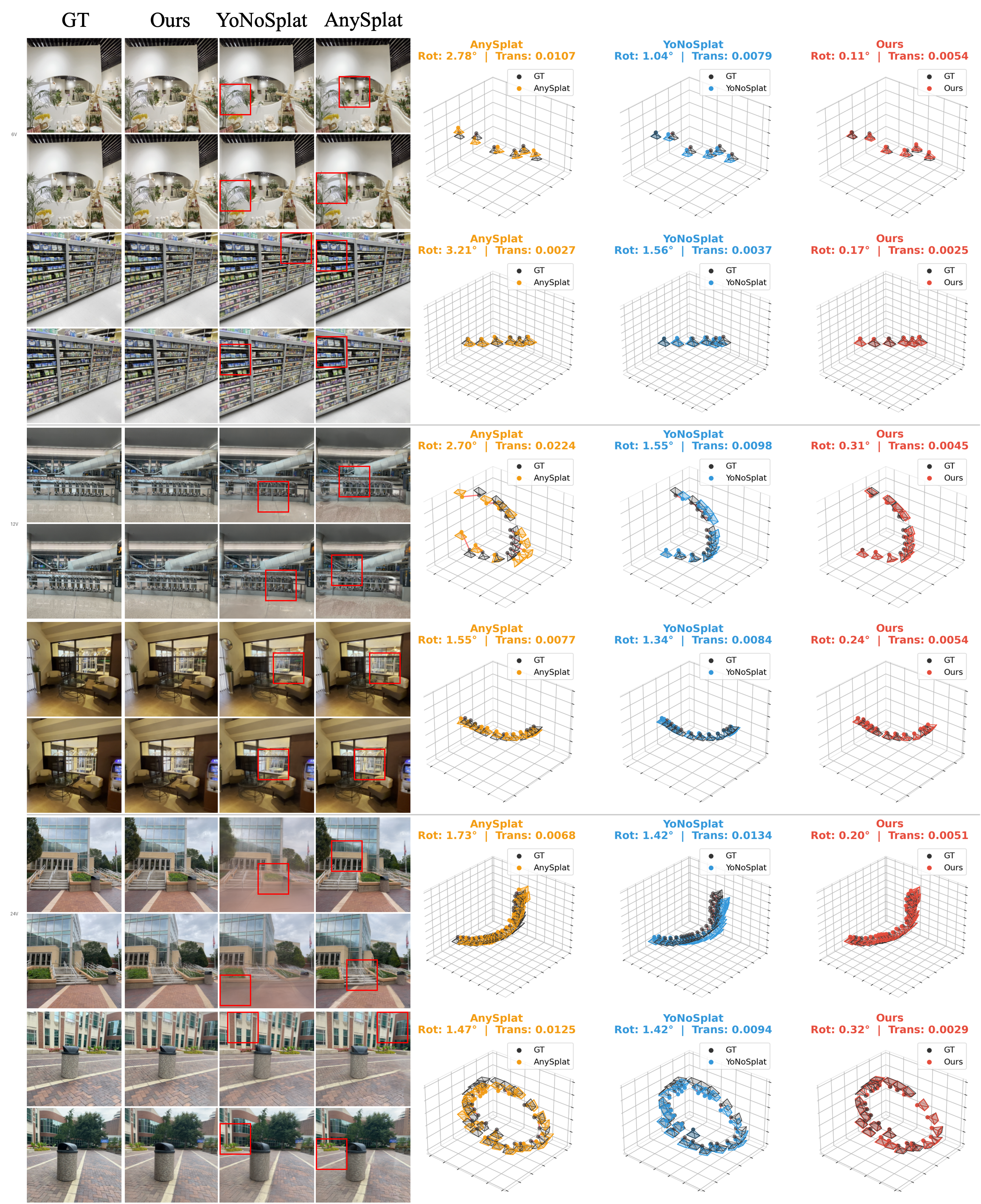}
  \caption{\textbf{Pose visualization and compared with representative methods on the DL3DV dataset.} Left: the rendering visualization of the target views. Right: we visualize the predicted context cameras, which shows our model can handle the pose drift problem.}
  \label{fig:supple_dl3dv_pose}
\end{figure}

\noindent {\bf More Visualizations on Pose Estimation.}
We provide more results for pose estimation in~\cref{fig:supple_dl3dv_pose}.

\begin{figure}[H]
  \centering
  \includegraphics[width=0.9\textwidth]{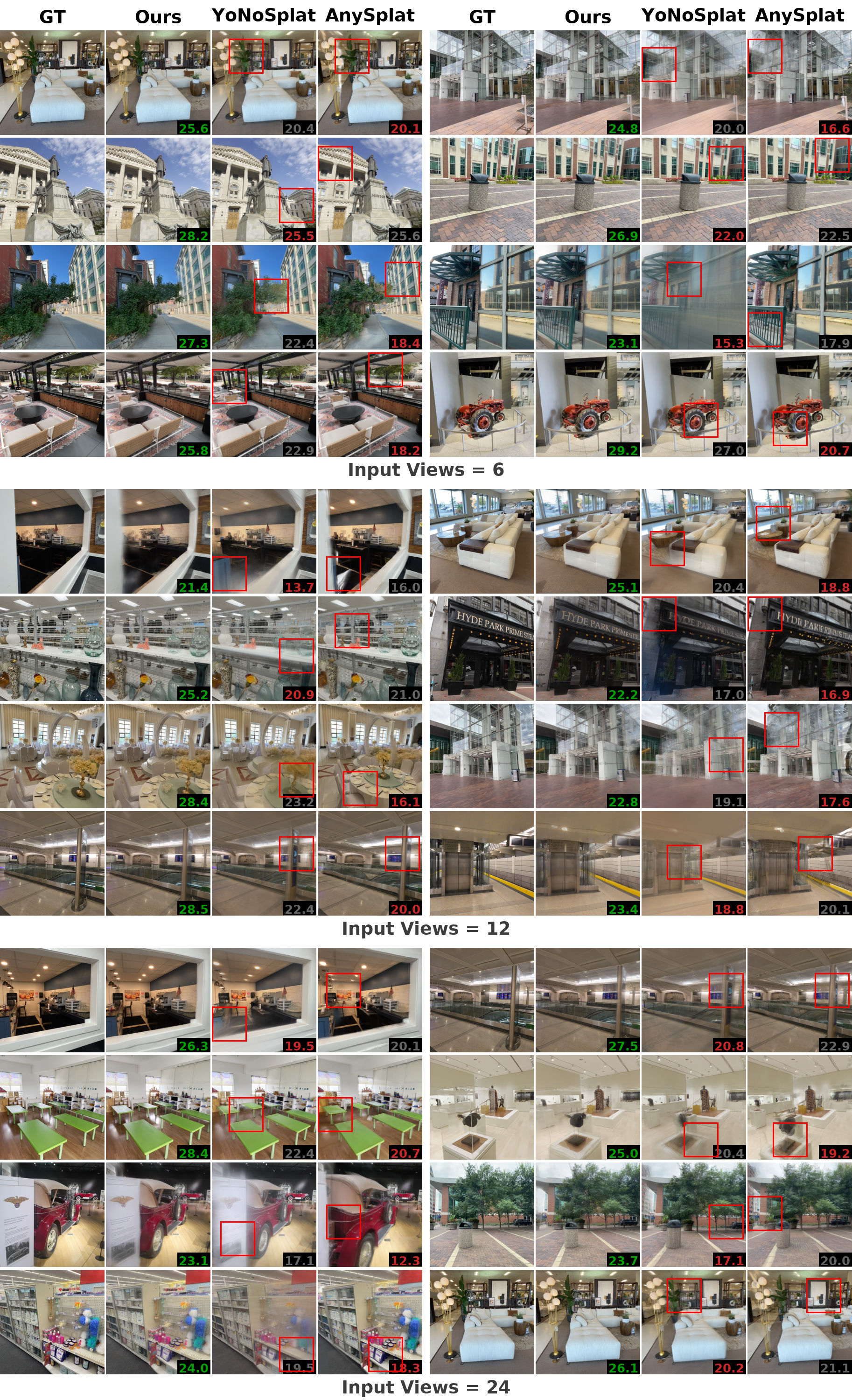}
  \caption{\textbf{More qualitative comparison of novel view synthesis on the DL3DV test set.} Here we report PSNR in the right corner in the pose-free, calibration-free setting. Our model produces higher-quality results compared to others. }
  \label{fig:supple_dl3dv_visualization}
\end{figure}

\noindent {\bf More Visualizations on Novel View Synthesis.}
We provide more novel view synthesis results in ~\cref{fig:supple_dl3dv_visualization}.

%
%

\end{document}